# SDE-DET: A Precision Network for Shatian Pomelo Detection in Complex Orchard Environments

Yihao Hu[a+], Pan Wang[a+], Xiaodong Bai[a*], Shijie Cai[b], Hang Wang[a], Huazhong Liu[a], Aiping Yang[c], Xiangxiang Li[c], Meiping Ding[c], Hongyan Liu[d], Jianguo Yao[e]

[a] School of Computer Science and Technology, Hainan University, Haikou 570228, Hainan, China.

[b] College of Computer and Control Engineering, Northeast Forestry University, Harbin 150040, China.

[c] Agricultural Meteorological Center, Jiangxi Meteorological Bureau, Nanchang 330045, Jiangxi, China.

[d] School of Breeding and Multiplication (Sanya Institute of Breeding and Multiplication), Hainan University, Haikou 570228, Hainan, China.

[e] School of Communications and Information Engineering, Nanjing University of Posts and Telecommunications, Nanjing 210003, Jiangsu, China.

[+]These authors contributed to the work equally.

*Correspondence: xiaodongbai@hainanu.edu.cn (X. Bai); Tel.: +86 0898 6627 1330.

**Abstract**

Pomelo detection is an essential process for their localization, automated robotic harvesting, and maturity analysis. However, detecting Shatian pomelo in complex orchard environments poses significant challenges, including multi-scale issues, obstructions from trunks and leaves, small object detection, etc. To address these issues, this study constructs a custom dataset STP-AgriData and proposes the SDE-DET model for Shatian pomelo detection. SDE-DET first utilizes the Star Block to effectively acquire high-dimensional information without increasing the computational overhead. Furthermore, the presented model adopts Deformable Attention in its backbone, to enhance its ability to detect pomelos under occluded conditions. Finally, multiple Efficient Multi-Scale Attention mechanisms are integrated into our model to reduce the computational overhead and extract deep visual representations, thereby improving the capacity for small object detection. In the experiment, we compared SDE-DET with the Yolo series and other mainstream detection models in Shatian pomelo detection. The presented SDE-DET model achieved scores of 0.883, 0.771, 0.838, 0.497, and 0.823 in Precision, Recall, mAP@0.5, mAP@0.5:0.95 and F1-score, respectively. SDE-DET has achieved state-of-the-art performance on the STP-AgriData dataset. Experiments indicate that the SDE-DET provides a reliable method for Shatian pomelo detection, laying the foundation for the further development of automatic harvest robots.

## 1.Introduction

Shatian Pomelo, a citrus fruit belonging to the Rutaceae family, is rich in nutritional value and provides various essential nutrients to the human body, which explains its widespread popularity [1]. It was first cultivated in Shatian Village, Rong County, Guangxi Province, P.R. China. Currently, pomelos have become one of the major agricultural products, widely planted in provinces such as Guangdong, Guangxi, and Jiangxi in China. Traditionally, the harvesting of Shatian pomelo is primarily manual. However, due to their large size and heavy weight, manual harvesting is time-consuming and labor-intensive, especially in large-scale cultivation [2]. To overcome these challenges and improve harvesting efficiency, researchers are increasingly focusing on developing automated harvesting technologies [3]. Among them, the detection of pomelos stands as the first and most crucial crop.

In smart agriculture, fruit detection has always been an important research direction, generating a large number of valuable research works [4-8]. In the early stages, traditional machine vision was applied to fruit detection, focusing on image processing, color analysis, feature extraction, and statistical modeling[9, 10]. The utilization of machine learning algorithms, like Support Vector Machines (SVM) and Random Forest (RF), led to notable enhancements in fruit detection. However, machine learning still faced limitations, including reliance on hand-crafted features and difficulties adapting to diverse environments [11]. These limitations prompted researchers to explore deep learning, which can automatically learn features from images, simplifying detection and significantly improving accuracy and adaptability.

Recently, with the advancement of technology, Artificial Intelligence has made significant progress and has been extensively applied in smart agriculture [12-18]. Indeed, these novel advancements in AI theory are particularly notable in their extensive application to fruit detection. Various deep learning methods for fruit detection have been proposed, mainly categorized into two-stage detection methods such as Faster R-CNN [19] and Cascade R-CNN [20], and single-stage detection algorithms like SSD [21], the Yolo series [22-24], CenterNet [25], and EfficientDet [26], etc. Among these single-stage methods, the Yolo series stands out due to its continuous improvements and superior performance.

In terms of two-stage models, Jia, et al. [27] proposed an optimized Mask R-CNN model that demonstrated excellent performance in apple overlap detection and segmentation, achieving a precision of 97.31% and a recall of 95.70%. Gao, et al. [28] proposed a multi-class apple detection method using Faster R-CNN to overcome previous limitations of treating all apples as a single class. This method identified apples in non-occluded, leaf-occluded, branch/wire-occluded, and fruit-occluded conditions, achieving average precisions of 0.909, 0.899, 0.858, and 0.848, respectively. López-Barrios, et al. [29] presented an enhanced Mask R-CNN model for detecting and segmenting green sweet pepper peduncles and fruits, achieving an 84.53% precision for fruits and a 71.78% precision for peduncles, significantly improving the performance in challenging greenhouse conditions. For single-stage models, Goyal, et al. [30] developed a Yolov5-based model for the detection and assessment of fruit quality. The initial stage focused on fruit detection, where the model demonstrated a mean Average Precision (mAP) of 92.80%. The subsequent stage was dedicated to assessing fruit quality, achieving a mAP of 99.60% for apples and 93.10% for bananas. Yang, et al. [31] put forward a novel strawberry ripening detection model LS-Yolov8s. Based on Yolov8s and incorporating the LW-Swin Transformer module, this model achieved a mAP of 93.8% on the test set. Chen, et al. [32] developed a Multi-Task Deep Convolutional Network, MTD-Yolov7, based on Yolov7 improvements to detect cherry tomatoes. This model enhanced detection accuracy by incorporating two decoders and utilizing SIoU instead of CIoU, resulting in a mAP of 86.6%. Zhu, et al.

[33] put forward Yolov5s-CEDB, an enhanced deep learning network for oil tea detection. This model enhanced local and global feature extraction by using coordinate attention mechanisms and deformable convolution, achieving a mAP of 91.4% and an F1-score of 89.6%. Lu, et al. [34] proposed a novel green citrus detection method that combined CNN with Transformer, reducing parameters by 9.6% and improving accuracy by 1.5% compared to Yolov5s. Yu, et al. [35] proposed MLG-YOLO for winter jujube detection and 3D localization, enhancing Yolov8n with 3.5% higher mAP, 61% fewer parameters, and precise positioning.

Despite the achievements of existing object detection models in fruit detection, challenges still remain in detecting Shatian pomelo. Firstly, the multi-scale issue poses a significant challenge in pomelo detection, as pomelos that are farther away appear smaller, while those in the foreground appear larger. Secondly, occlusion by leaves and trunks is common in our dataset due to the large size of the Shatian pomelo, making accurate detection especially difficult. Moreover, small object detection poses an additional challenge in our study which has not been thoroughly resolved in previous research. Pomelos that are farther from the camera appear very small in the image, leading to a serious loss of their feature information and making their detection very complicated. Furthermore, the texture and color of Shatian pomelos are similar to leaves and trunks, making it difficult to distinguish them from the background. To effectively address these challenges, this study presents an automatic detection method for Shatian pomelo based on the Yolov8n model. As a lightweight variant within the Yolov8 family, Yolov8n is optimized for efficient operation in resource-constrained conditions, making it well-suited for the Shatian pomelo detecting and harvesting robot. The main contributions of this paper include:

1. We present STP-AgriData, a new dataset for Shatian pomelo detection, establishing a robust foundation for future research. To our knowledge, this is the first dataset that combines field-collected and publicly sourced data, thereby enhancing the diversity and comprehensiveness of the Shatian pomelo dataset.

2. We propose a new network SDE-DET, designed to achieve automatic, contactless, and accurate detection of Shatian pomelo in complex orchard environments. In SDE-DET, the Star Block, Deformable Attention, and Efficient Multi-Scale Attention mechanisms are adopted to reduce feature loss, preserve key information from the original image, and enhance the detection of small and occluded Shatian pomelos.

3. Experimental results show that SDE-DET outperforms the state-of-the-art on the STP-AgriData dataset, achieving scores of 0.883, 0.771, 0.838, 0.497 and 0.823 in Precision, Recall, mAP@0.5, mAP@0.5:0.95 and F1-score, respectively.

The rest of the paper is structured as follows: Section 2 introduces image acquisition and dataset creation, and Section 3 presents our model's structure and improvements. Section 4 evaluates the improved YOLOv8n by analyzing the results of comparative and ablation experiments, and Section 5 summarizes the research work.

## 2. The Shatian pomelo dataset
### 2.1. Data acquisition and analysis

As shown in Fig. 1(a), this study's dataset was collected from Taipingshan Village,

Ganzhou City, Jiangxi Province, China, with geographical coordinates of 114°51′45″ East longitude and 25°20′52″ North latitude. The elevation of the area was 193.1 meters. The total covered area was approximately 49.45 mu.

To ensure the rigor and reproducibility of field data collection, we have supplemented key details regarding the environmental conditions during photography and the sampling methodology for pomelo trees—two aspects critical to clarifying the data context.

First, regarding environmental conditions: All field images were captured under natural light to reflect real-world growth scenarios of pomelo trees. Photography was conducted intermittently when time permitted, with most sessions taking place in the afternoon; this timing was chosen to avoid the overexposure of midday sun or insufficient light in the early morning and evening. Additionally, all data collection was carried out on sunny days, eliminating interference from weather factors such as rain, fog, or overcast skies that could affect image clarity. The entire field sampling period occurred in winter, and specific shooting dates are recorded in the photo metadata for easy traceability. During this period, the pomelo fruits were in the stage from enlargement to maturity, which is a typical and representative developmental phase for Shatian pomelo in this regional climate.

Second, in terms of sampling methodology: To ensure uniformity and comparability across samples, we implemented strict selection criteria for pomelo trees. Sampling was exclusively conducted on mature pomelo trees aged 5–7 years, as this age group represents stable fruiting performance. Trees with a height of approximately 2 meters were selected to maintain consistency in shooting height and perspective, and images were captured from the outer south side of the tree crown—this position ensures sufficient light exposure and captures a representative distribution of pomelo fruits. Furthermore, all sampled trees were subject to identical agricultural management practices: organic fertilizers were used to provide nutrients, and irrigation was conducted via an integrated water-fertilizer facility to ensure balanced water and nutrient supply. No special treatments (such as additional pesticide application or pruning) were applied to the sampled trees, which guarantees uniformity in water, fertilizer, and pruning conditions across all experimental subjects.

Fig. 1(b) shows our multiple real-time monitoring instruments for data collection, including the Hikvision DS-8632N-I8 model. Fig. 1(c) illustrates the working principle of image acquisition equipment, which involves the process from image capture by the camera to data transmission via the wireless bridge and other network hardware. The collected images are shown in Fig. 1(d).

Regarding the camera settings for image capture, we would like to provide additional details to address previously noted information insufficiencies. During data collection, core camera parameters such as aperture, shutter speed, and ISO were configured in accordance with the camera's factory default settings, and this specification will be clearly added to the revised manuscript. Furthermore, other relevant camera setup details are supplemented as follows: the image resolution was set to 1920×1080 pixels with the file format in JPEG; the horizontal distance between the camera and the pomelo trees was controlled within the range of 3 to 5 meters, while the camera height was fixed at 2 meters.

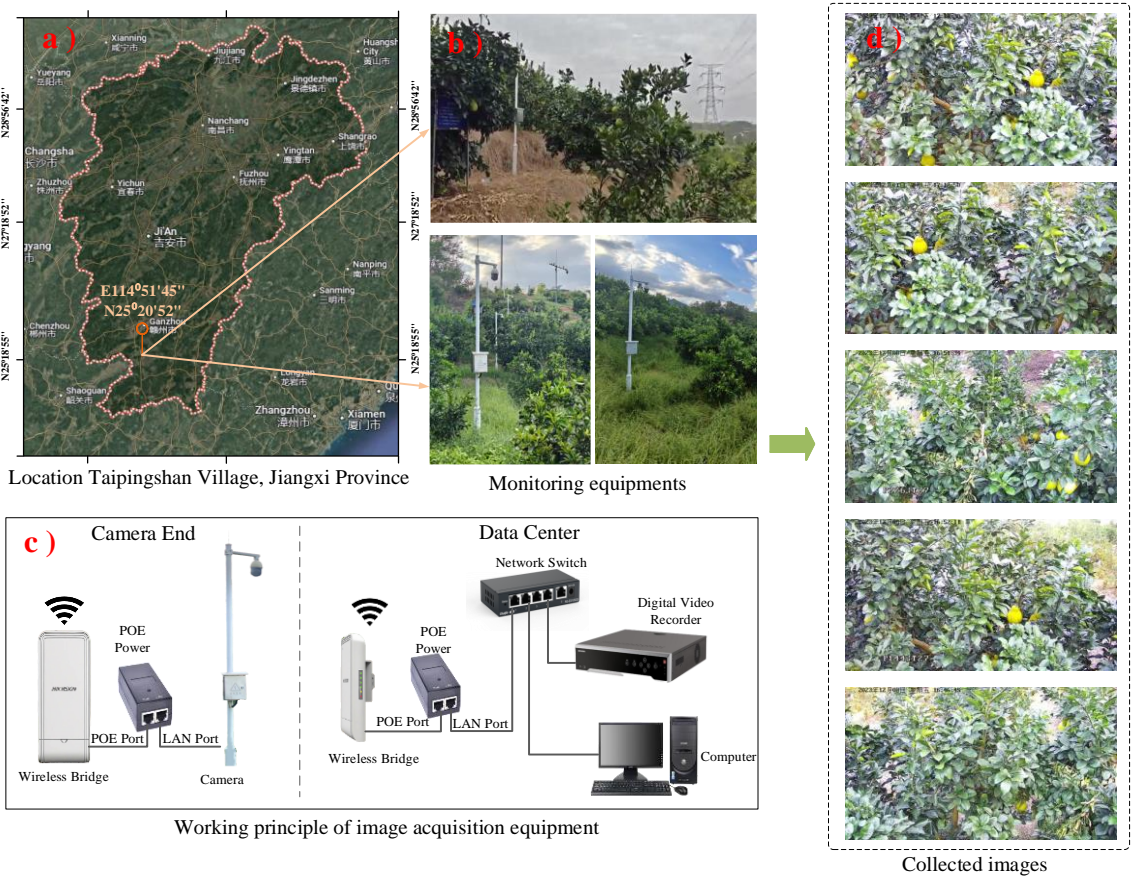

**Fig. 1.** Data acquisition

Additionally, to further improve the diversity and comprehensiveness of the Shatian pomelo image data, we gathered relevant data from publicly available online sources, as shown in the 3rd and 4th rows of Fig. 2. By combining field-collected and publicly sourced data, we were able to collect images of Shatian pomelo more comprehensively from different regions and environments. This data collection method not only enhanced the richness and representativeness of the dataset but also boosted the model's accuracy and generalization capabilities. Consequently, the research outcomes achieved higher detection precision and greater practical application value.

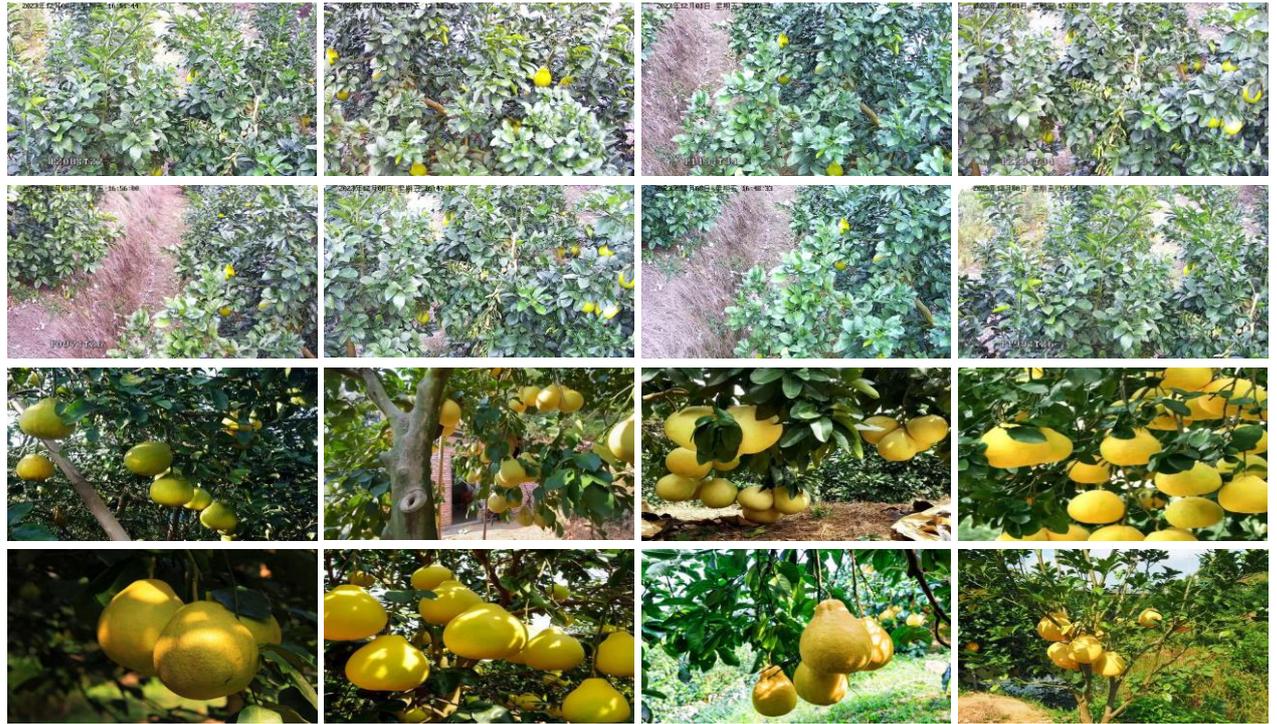

**Fig. 2.** Partial images of the dataset

### 2.2. Data preprocessing

During the data preprocessing stage, we performed the following steps to ensure the quality and diversity of the dataset. First, we consolidated all images and annotation files into a unified directory, ensuring each image corresponds to its respective annotation file. Subsequently, we cleaned the data by removing one damaged image file, resulting in a final set of 317 images of Shatian pomelo, which formed the STP-AgriData dataset. Next, we split the dataset into training and testing sets in a 6:4 ratio using a fixed random seed to ensure the randomness and reproducibility of the split. As a result, the training set included 190 images, and the testing set included 127 images. To enhance our model's robustness and generalization capability, we applied data augmentation to 190 Shatian pomelo images. Specifically, we used six methods: brightness adjustment, contrast adjustment, denoising, grayscale transformation, horizontal flipping, and vertical flipping, as shown in Fig. 3. Through this augmentation, we increased the total number of images to 1,330. These preprocessing steps significantly improved the quality and diversity of the dataset, providing a solid foundation for model training and evaluation, and ultimately enhancing the model's generalization ability.

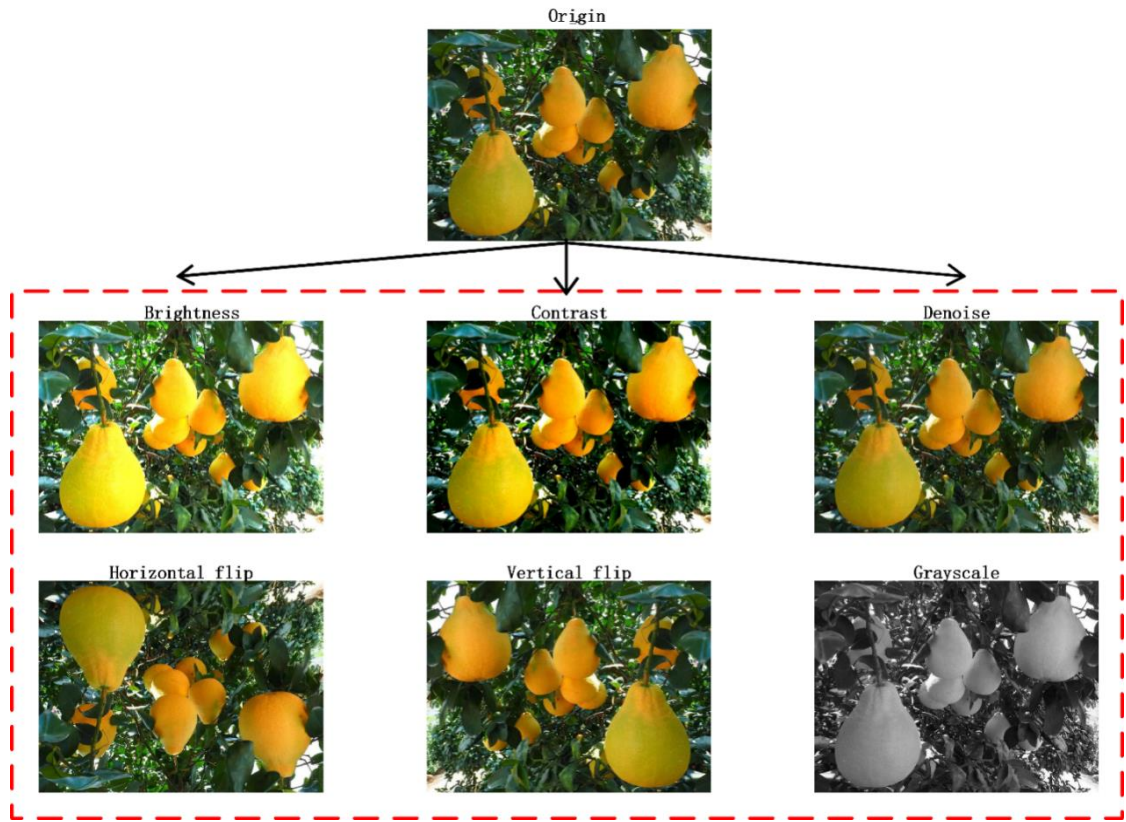

**Fig. 3.** Data augmentation

## 3. SDE-DET Network for Shatian Pomelo Detection

In this section, the proposed network architecture is introduced. As illustrated in Fig. 4, we constructed the SDE-DET Network by improving the Yolov8n baseline model with the Star Block and other advanced deep learning methods; as shown in Table 1, the parameters of the SDE-DET Network are detailed. The Network incorporated three key improvements. First, by utilizing the Star Block, we aimed to minimize the feature loss during convolution while significantly increasing the computational load, effectively preserving high-dimensional feature information. Then, we integrated Deformable Attention into our model's backbone, allowing the model to focus more effectively on occluded Shatian pomelos. Finally, we adopted multiple Efficient Multi-Scale Attention Mechanisms in our Network to enhance its ability to learn relevant features, thus achieving more effective feature fusion and improving its ability to detect Shatian pomelos of different sizes. The Network parameters are presented in Table 1. These improvements significantly enhanced the network's detection performance in complex environments, demonstrating higher accuracy and robustness in the Shatian pomelo detection task. The source code of our presented SDE-DET will be shared from https://github.com/mistletoe111/SDE-DET.

Table 1. SDE-DET network parameters

| Operation | Input | Output | Step | Kernel |
|---|---|---|---|---|
| ConvModule | (640,640,3) | (320,320,8) | 2 | (3, 3) |
| Star Block | (320,320,8) | (320,320,8) | | |
| ConvModule | (320,320,32) | (160,160,32) | 2 | (3, 3) |
| ConvModule | (160,160,32) | (80,80,64) | 2 | (3, 3) |
| CSPLayer_2Conv | (80,80,64) | (80,80,64) | 1 | (1, 1) |
| Feat1 | | (80,80,64) | | |
| ConvModule | (80,80,64) | (40,40,128) | 2 | (3, 3) |
| Feat2 | | (40,40,128) | | |
| ConvModule | (40,40,128) | (20,20,256) | 2 | (3, 3) |
| CSPLayer_2Conv | (20,20,256) | (20,20,256) | 1 | (1, 1) |
| SPPF | (20,20,256) | (20,20,256) | | |
| Deformable Attention | (20,20,256) | (20,20,256) | | |
| Feat3 | | (20,20,256) | | |

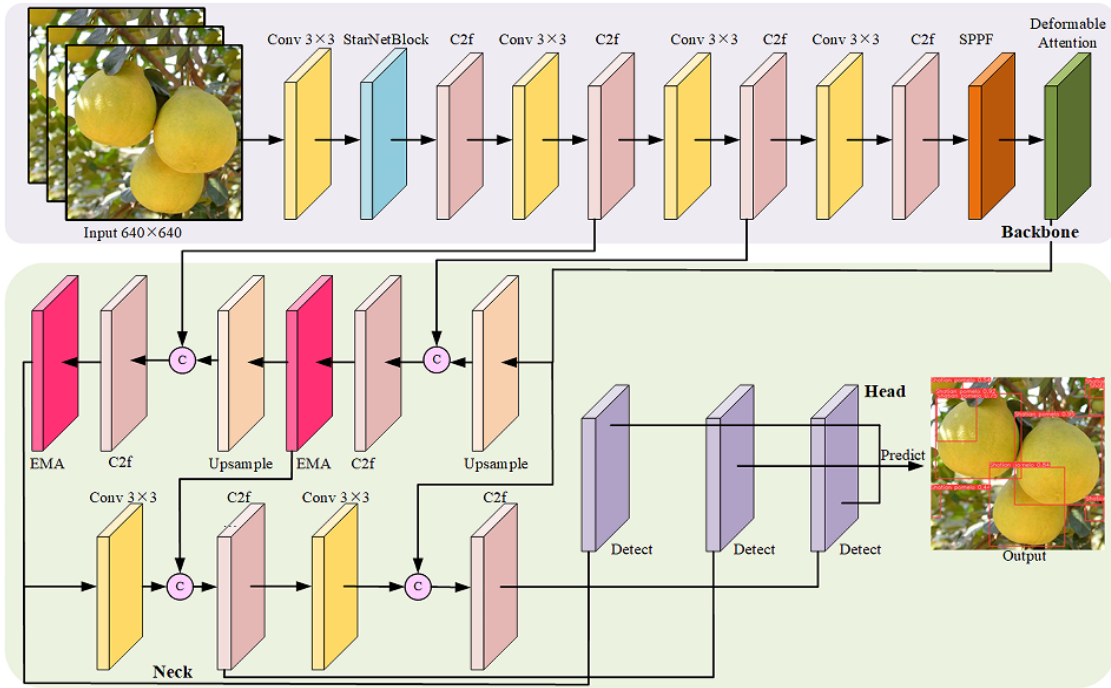

Fig. 4. SDE-DET network

### 3.1. Utilization of the Star Block

During our comprehensive analysis of the Yolov8 backbone network, we identified a critical issue related to the initial stages of the network architecture. Specifically, we observed that the dimensional gap between the first convolutional layer and the input image was excessively large. This discrepancy results in a significant loss of feature information during early convolutional operations, which compromises the ability of the network to accurately capture the detailed characteristics of the input images. To address this issue, we adopted the Star Block from StarNet proposed by Ma, et al. [32], whose structure is depicted in Fig. 5. The core of the Star Block is the redefined Star operation:

$$w_1^T y * w_2^T y = \left(\sum_{i=1}^{d+1} w_1^i y^i\right)\left(\sum_{j=1}^{d+1} w_2^j y^j\right) \quad (1)$$

$$\sum_{i=1}^{d+1} \sum_{j=i}^{d+1} \gamma(i,j) y^i y^j \quad (2)$$

where γ(i,j) can be described as:

$$\gamma(i,j) \begin{cases} w_1^i w_2^j & if\ i == j \\ w_1^i w_2^j + w_1^j w_2^j & if\ i! = j \end{cases} \quad (3)$$

We define $w_1$, $w_2$, and $y$ to each be in $\mathbb{R}^{(d+1)\times 1}$, where d represents the number of input channels.

The Star operation performs pairwise feature multiplications across different channels, similar to kernel functions. In deep neural networks with multiple stacked layers, it achieves exponential growth in the implicit dimensionality within a compact feature space. This leads to virtually infinite dimensions while retaining the advantages of high-dimensional representations. Therefore, the Star Block can map the input into a high-dimensional, nonlinear feature space, which not only avoids increasing the computational load but also effectively preserves the feature information in the image. Consequently, we replaced the original convolutional block with 16 channels. Instead, we used an original convolutional block with 8 channels combined with a Star Block followed by a convolutional layer with 32 channels. These improvements enable the effective extraction and preservation of high-dimensional features, resulting in a more accurate detection of Shatian pomelo.

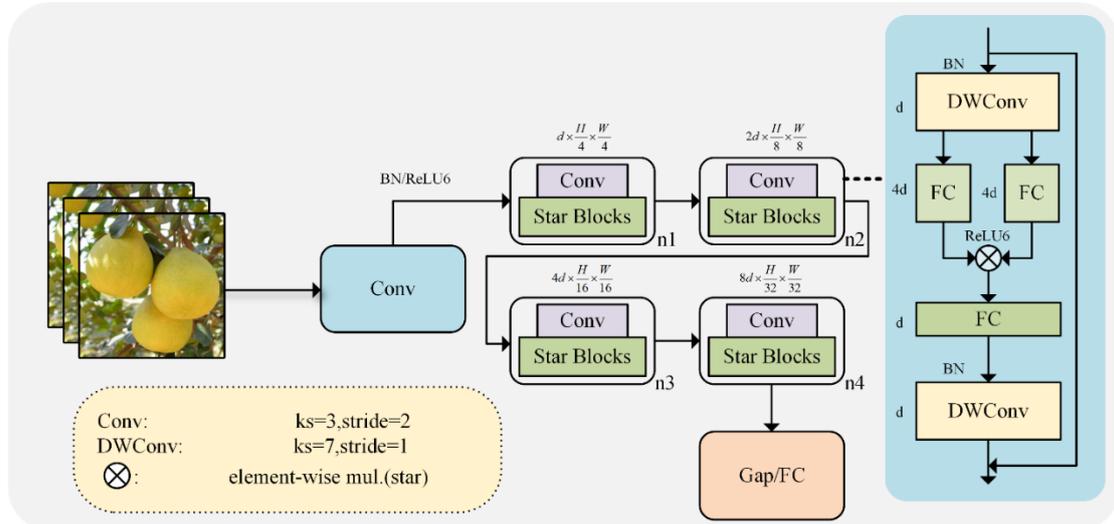

**Fig. 5.** The StarNet and Star Block

**3.2. Adoption of the Deformable Attention**

Deformable Attention is an innovative attention module designed to enhance a model's ability to handle local and translation variations [36]. It enables the model to focus more effectively on critical areas when processing data, such as images or videos. The core idea behind this module lies in its ability to dynamically adjust the attention focus, allowing the model to capture important information from the input data more accurately.

As shown in Fig. 6, Deformable Attention first creates a uniformly distributed reference point grid based on the feature map. These reference points serve as the initial attention positions. To compute the offsets for each reference point, a linear projection is first applied to the feature map to obtain the query embeddings $Q$ using the query projection matrix $W_q$. These query embeddings are then passed through a subnetwork β_offset (·), which generates the corresponding offsets.

$$Q = W_q x, \widetilde{K} = W_k \tilde{x}, \widetilde{V} = W_v \tilde{x} \tag{4}$$

$$\Delta p = \beta_{offset}(Q), \tilde{x} = \varphi(x; p + \Delta p) \tag{5}$$

Here, $\widetilde{K}$ and $\widetilde{V}$ represent the deformed key and value embedding respectively. The term $\varphi(x; p + \Delta p)$ denotes the bilinear interpolation function applied to the input feature map at the adjusted positions $p + \Delta p$, resulting in the deformed features $\tilde{x}$.

Finally, the Deformable Attention module aggregates information using these deformed key-value pairs according to the standard multi-head attention mechanism. A multi-head self-attention (MHSA) block with M heads can be expressed as:

$$Q = W_q x, K = W_K x, V = W_v x \tag{6}$$

$$h^{(m)} = \alpha \left( \frac{q^{(m)} k^{(m)\top}}{\sqrt{d}} \right) v^{(m)}, m = 1, \ldots, M \tag{7}$$

$$h = Concat(h^{(1)}, \ldots, h^{(M)}) W_o \tag{8}$$

α(·) represents the softmax function, and $h^{(m)}$ refers to the embedding output from the m-th attention head. The embeddings for query, key, and value are designated as $q^{(m)}, k^{(m)}, v^{(m)}$ respectively. The corresponding projection matrices are labeled as $W_q, W_K, W_v, W_o$.

Through the multi-head attention mechanism, the model integrates information from various representation subspaces, thus generating more comprehensive and distinctive feature representations. In this way, Deformable Attention not only enhances the model's representational capabilities but also improves its adaptability and robustness to variations in different input data.

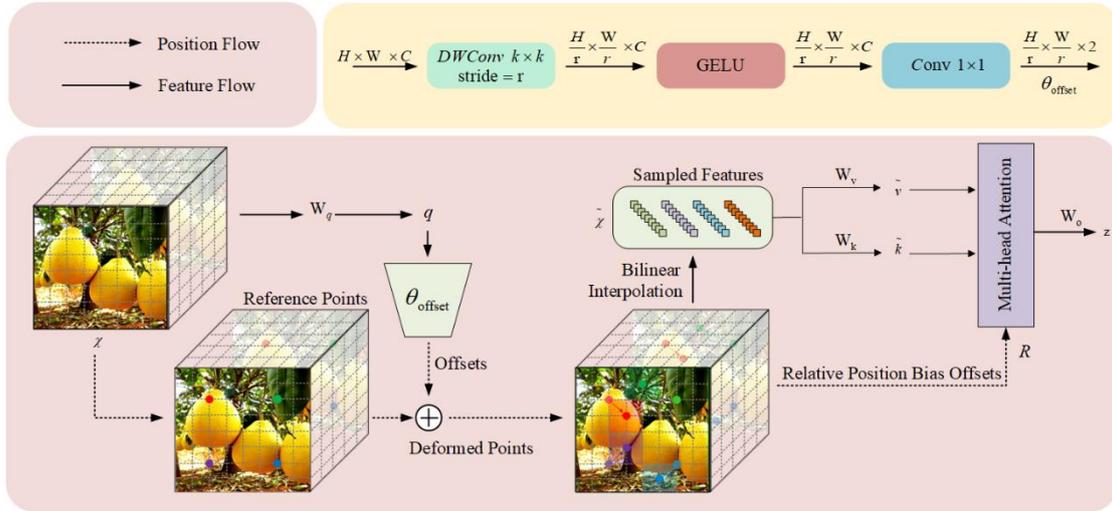

**Fig. 6.** Deformable Attention

**3.3. Application of Efficient Multi-Scale Attention Mechanism**

The Efficient Multi-Scale Attention Mechanism, proposed by Ouyang, et al. [37], is an efficient multi-scale attention mechanism that focuses on preserving information across each channel while reducing the computational overhead, as illustrated in Fig. 7. This is achieved by re-aligning some channels to the batch dimension and grouping the channel dimension into multiple sub-features, enabling the learning of different semantics. The grouping can be represented by $X = [X_0, X_i, \ldots, X_{G-1}]$, $X_i \in R^{C//G \times H \times W}$. In addition, the Efficient Multi-Scale Attention Mechanism utilizes cross-space information aggregation across different spatial dimensions to achieve more comprehensive feature aggregation. Specifically, it uses two-dimensional (2D) global average pooling to process the output of a $1 \times 1$ branch, thereby encoding global spatial information. The 2D global pooling operation is expressed as

$$o_c = \frac{1}{H \times W} \sum_{j}^{H} \sum_{i}^{W} x_c(i,j) \qquad (9)$$

This operation encodes global information and models long-range dependencies, allowing the Efficient Multi-Scale Attention mechanism to avoid the extensive sequential processing and deeper network depth typically found in traditional attention mechanisms. It effectively learns channel descriptions without compressing channel dimensions. In addition, it generates superior pixel-level attention for high-level feature maps. Given the strong capability of the Efficient Multi-Scale Attention mechanism to extract deep visual representations, we adopted multiple Efficient Multi-Scale Attention mechanisms for our model. This improvement allows the model to capture and learn relevant features better, thereby enhancing the overall performance and accuracy.

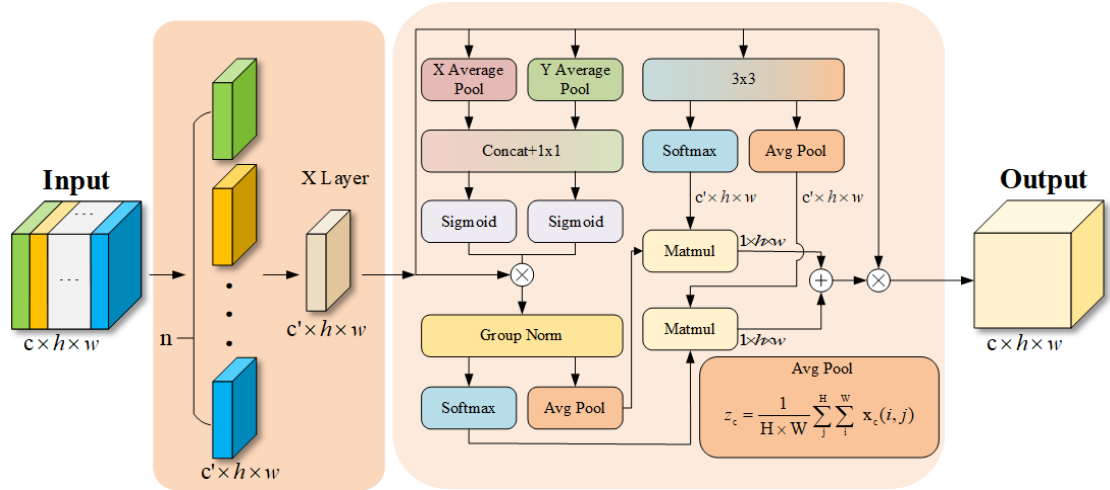

**Fig. 7.** Efficient Multi-Scale Attention Mechanism

## 4 Results and discussions

### 4.1 Experimental platforms and hyperparameters

The operating system used for the experiments in this paper is Linux. The host CPU is an AMD EPYC 7542, and the host GPU is an NVIDIA GeForce GTX 4090. The deep learning framework used is PyTorch 2.4.0, which is paired with CUDA 12.1, and the Python version is 3.11.5. The batch size is set to 8, and the preprocessed images were uniformly scaled to $640 \times 640$ before being fed into the model. Based on our previous experiments, the optimal model usually appears within the first 100 epochs of training. Therefore, we set the number of training epochs to 100. Other parameters, such as the initial learning rate and weight decay, are left at their default values.

### 4.2 Evaluation metrics

Object detection is an important task in computer vision that involves identifying the location and class of individual objects. Various evaluation metrics need to be used to evaluate the accuracy and efficiency of object detection methods. This paper describes the commonly used evaluation metrics for object detection.

#### 4.2.1 Precision and Recall

Precision and Recall are the fundamental evaluation metrics for object detection. Precision is defined as the ratio of correct samples in the test results to the total number of samples in the test results. The calculation formula is shown in (10), where TP stands for true positive, which is the number of samples correctly detected as positive, and FP stands for false positive, which is the number of samples incorrectly detected as positive.

$$\text{Precision} = \frac{TP}{TP + FP} \qquad (10)$$

Recall is defined as the ratio of correct samples in the test results to the total number of true samples. It reflects the completeness of the test results. The calculation formula is shown in (11). FN denotes false negatives, which means the number of positive samples that are not detected by the model. It is worth noting that precision and recall are often contradictory, meaning that a detector may increase accuracy while decreasing recall.

$$\text{Recall} = \frac{TP}{TP + FN} \qquad (11)$$

### 4.2.2 Mean Average Precision

Mean Average Precision (mAP) is a key evaluation metric in object detection that assesses a model's overall detection accuracy across different classes. To calculate mAP, we first compute the Average Precision (AP) for each class by integrating the Precision-Recall (P-R) curve:

$$\text{AP} = \int_0^1 p(r)dr \tag{12}$$

where p(r) represents the precision as a function of recall. The mAP is then obtained by averaging the AP values of all classes:

$$\text{mAP} = \frac{1}{N}\sum_{i=1}^{N} \text{AP}_i \tag{13}$$

with N is the total number of classes and $\text{AP}_i$ is the average precision of the i-th class. A higher mAP value indicates a better overall detection performance of the model.

### 4.2.3 IoU

The intersection over union (IoU) quantifies the overlap between the predicted bounding box and the ground truth box in object detection. It is calculated as:

$$\text{IoU} = \frac{\text{Area of Intersection}}{\text{Area of Union}} \tag{14}$$

where the Area of Intersection is the overlapping area between the predicted and actual boxes, and the Area of Union is the total area covered by both boxes. A higher IoU value indicates better localization accuracy, reflecting the ability of the model to precisely predict object boundaries.

### 4.2.4 F1 score

The F1 score combines precision and recall into a single metric to evaluate the detection effectiveness of the model. It is calculated as the harmonic mean of the precision and recall:

$$F1 = 2 \cdot \frac{precision \cdot recall}{precision + recall} \tag{15}$$

A higher F1 score indicates a better overall detection performance of the model.

### 4.3. Comparison experiments

This study adopted a series of new enhancements to the original Yolov8n model, aiming to improve the model's accuracy and efficiency. In this study, we compared it with the Yolo family and other mainstream models to experimentally evaluate the performance of SDE-DET. The experiments were conducted under consistent hardware and software environments to follow the principle of control variables and ensure the reproducibility and fairness of the experimental results.

Table 2 compares the performance of the SDE-DET with those of Yolov8, Yolov9, and Yolov10. Compared to Yolov8n and Yolov8s, mAP@0.5 of SDE-DET improved by 5.1% and 2.9%, respectively. Compared to Yolov9c, Yolov10n, and Yolov10s, mAP@0.5 of SDE-DET improved by 3.4%, 7.3%, and 5.9%, respectively. Additionally, SDE-DET outperformed the Yolo9 and Yolov10 series models across four metrics: precision, recall, mAP@0.5:0.95, and F1 score. This demonstrates that the proposed improvement is effective in terms of both recognition accuracy and computational efficiency for Shatian pomelo detection.

**Table 2.** Comparative experimental results of Yolo series models

| | Model | Precision | Recall | mAP@0.5 | mAP@0.5:0.95 | F1-Score |
|---|---|---|---|---|---|---|
| 2023 | Yolov8n | 0.862 | 0.700 | 0.787 | 0.458 | 0.773 |
| 2023 | Yolov8s | 0.853 | 0.745 | 0.809 | 0.467 | 0.796 |
| 2024 | Yolov9c [38] | 0.870 | 0.720 | 0.804 | 0.475 | 0.788 |
| 2024 | Yolov10n [39] | 0.851 | 0.673 | 0.765 | 0.442 | 0.752 |
| 2024 | Yolov10s [39] | 0.813 | 0.725 | 0.779 | 0.446 | 0.766 |
| 2024 | SDE-DET | **0.883** | **0.771** | **0.838** | **0.497** | **0.823** |

Table 3 shows the performance comparison of the SDE-DET with other mainstream models. Compared to Faster R-CNN, Cascade R-CNN, CenterNet, RTMDet-m, DDQ-4scale, DINO-4scale, and RT-DETR, SDE-DET's mAP@0.5 improves by 25.7%, 22.0%, 15.7%, 8.9%, 12.1%, 9.8%, and 5.1%, respectively. In terms of mAP@0.5:0.95, the improvements are 21.7%, 13.8%, 12.4%, 6.7%, 9.7%, 8.3%, and 4.0% respectively. In addition, SDE-DET achieves higher recognition rates than DDQ-4scale and RT-DETR. This indicates that the proposed improvement guarantees high recognition rates while maintaining low computational overhead in Shatian pomelo detection.

**Table 3.** Experimental results comparing SDE-DET with other models

| Proposed year | Model | mAP@0.5 | mAP@0.5:0.95 | Parameters(M) |
|---|---|---|---|---|
| 2017 | Faster R-CNN [19] | 0.581 | 0.280 | 40.34 |
| 2018 | Cascade R-CNN [20] | 0.618 | 0.359 | 88.06 |
| 2019 | CenterNet [25] | 0.681 | 0.373 | 32.24 |
| 2022 | RTMDet-m [40] | 0.749 | 0.430 | 24.70 |
| 2023 | DINO-4scale [41] | 0.717 | 0.400 | 48.03 |
| 2023 | DDQ-4scale [42] | 0.740 | 0.414 | 47.21 |
| 2024 | RT-DETR [43] | 0.787 | 0.459 | 42.76 |
| 2024 | SDE-DET | **0.838** | **0.497** | **3.29** |

These substantial performance gains can be attributed to the combined effect of the Star Block, Deformable Attention, and Efficient Multi-Scale Attention mechanism integrated into the SDE-DET model. The Star Block enables the model to capture details and objects in complex backgrounds better by introducing more hierarchical features. Deformable Attention enhances the recognition of the Shatian pomelo in occlusion situations by allowing the model to focus more effectively on critical areas. The Efficient Multi-Scale Attention Mechanism improves the detection of Shatian pomelos in complex environments by extracting depth visual representations, thus improving the detection of small-scale Shatian pomelos.

### 4.4 Ablation experiments

Ablation experiments can deepen the understanding of the interactions between the components of an object detection algorithm, determine the optimal model parameters, and improve the performance of the model. In addition, ablation experiments can identify the algorithm's limitations, offering researchers clear directions and methods for further optimization. This is evidenced by the results of the ablation experiments in Table 4, where the model metrics were significantly improved with the addition of the Star Block, Deformable Attention, and Efficient Multi-Scale Attention. Among them, the SDE-DET model performed the best in terms of precision, recall, mAP@0.5, mAP@0.5:0.95, and F1 values. It reached 0.883, 0.771, 0.838, 0.497, and 0.823, respectively, while maintaining a reasonable computational cost. It can also be seen that SDE-DET outperforms the baseline model by 2.1%, 7.7%, 5.1%, 3.9%, and 5.0% in precision, recall, mAP@0.5, mAP@0.5:0.95 and F1 values, respectively, which gives it a big advantage in the complex task of Shatian pomelo detection.

**Table 4.** Results of ablation experiments

| | +Deformable Attention | +Star Block | +EMA | Precision | Recall | mAP@0.5 | mAP@0.5:0.95 | F1-score |
|---|---|---|---|---|---|---|---|---|
| 0 | | | | 0.862 | 0.700 | 0.787 | 0.458 | 0.773 |
| 1 | √ | | | 0.870 | 0.716 | 0.794 | 0.456 | 0.786 |
| 2 | √ | √ | | 0.875 | 0.738 | 0.817 | 0.487 | 0.800 |
| 3 | √ | √ | √ | **0.883** | **0.771** | **0.838** | **0.497** | **0.823** |

As illustrated in Fig. 8, the curves of mAP@0.5 and mAP@0.5:0.95 for SDE-DET, the original Yolov8n, and related models are shown over 100 training iterations. The experiment involved 100 training rounds. As shown in Fig. 7, the SDE-DET model demonstrates a notable and consistent improvement in performance, particularly during the final 50 rounds of training. This sustained improvement indicates the robustness of the model against overfitting and highlights its strong learning capacity. In contrast, the original Yolov8n and other models exhibit signs of overfitting, indicating limitations in their ability to generalize effectively. This superior performance highlights the enhanced detection accuracy and robust generalization capabilities of the SDE-DET, underscoring the efficacy of the improvements made to the model architecture and its potential for reliable and precise object detection.

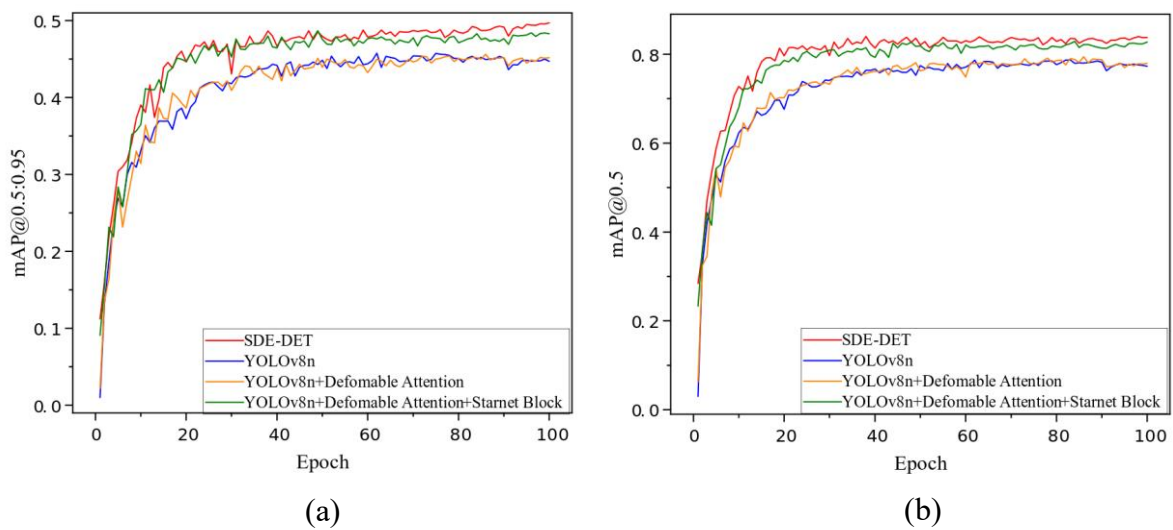

**Fig. 8.** (a) Curves of mAP@0.5 for each improved version with training iterations and (b) Curves of mAP@0.5:0.95 for each improved version with training iterations.

### 4.5 Visualization of network performance

Visualization provides a more intuitive view of the effectiveness of a model's detection. It allows information such as the location and size of objects detected by the model to be observed and can be used to assess the accuracy and efficiency of the model as well as to optimize the model.

Fig. 9 displays the partial detection results, where (1) indicates SDE-DET, and (2) indicates Yolov8n. In Fig. a1, a2, d1, and d2, the small pomelos in the upper right corner were not detected by Yolov8n, which shows the shortcomings of the baseline in detecting small objects. In Fig. b1, b2, e1, and e2, strong lighting makes it difficult to detect small targets. In Fig. c1 and c2, the original model resulted in false detections owing to the influence of strong lighting. In Fig. g1 and g2, because the color of the leaves was similar to that of Shatian pomelo, the original model mistook the leaves for Shatian pomelo during detection. In Fig. f1, f2, h1, and h2, it can be clearly observed that when the Shatian pomelos were densely distributed, Yolov8n struggled to achieve non-overlapping detection, indicating deficiencies in its feature-extraction capabilities.

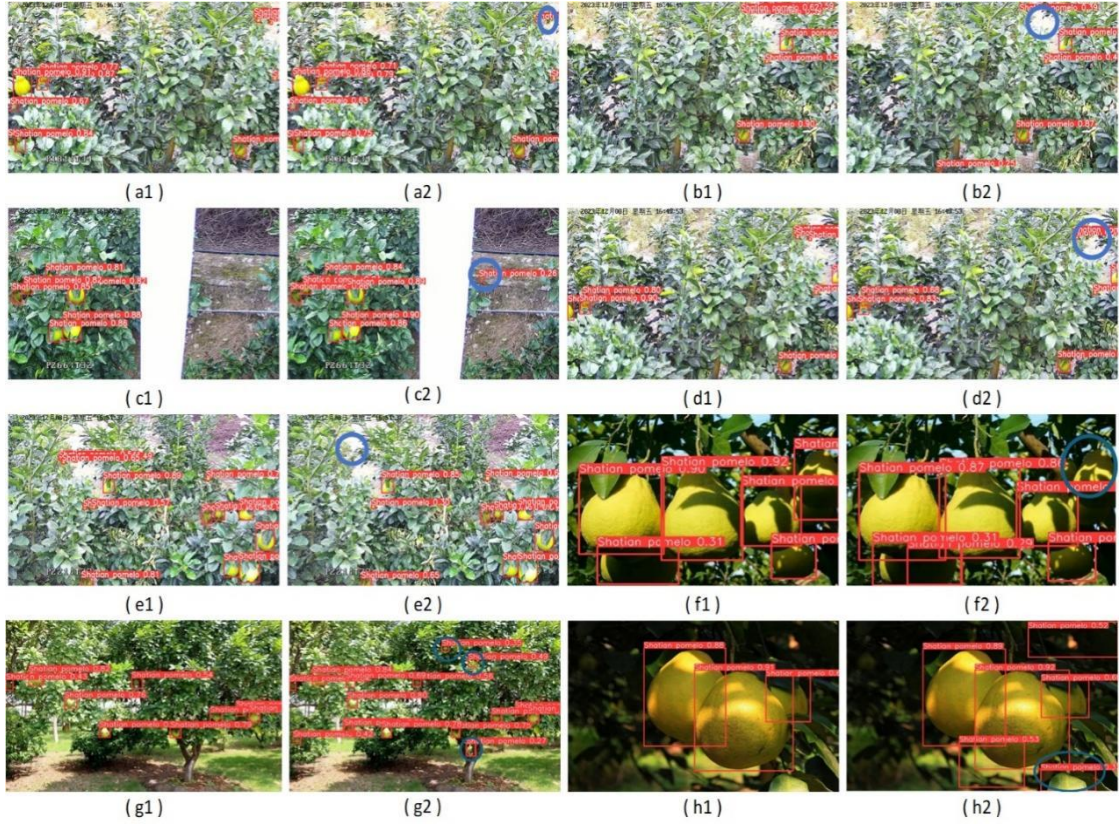

**Fig. 9. Comparison of SDE-DET and Yolov8n detection results**

The heat map generated by Grad-CAM highlights the significance of Shatian pomelo edge information. It can also reflect the extent of the contribution from color features. Fig. 10 presents some of the results, demonstrating the effectiveness of the model in capturing these features. In Fig. 10, the red color indicates the location and intensity of the object, with a higher intensity signifying greater confidence in the model's detection results. This visualization helps confirm that each Shatian pomelo is accurately localized and that the model has a high level of certainty in its detection, thereby validating the robustness and reliability of the model's performance.

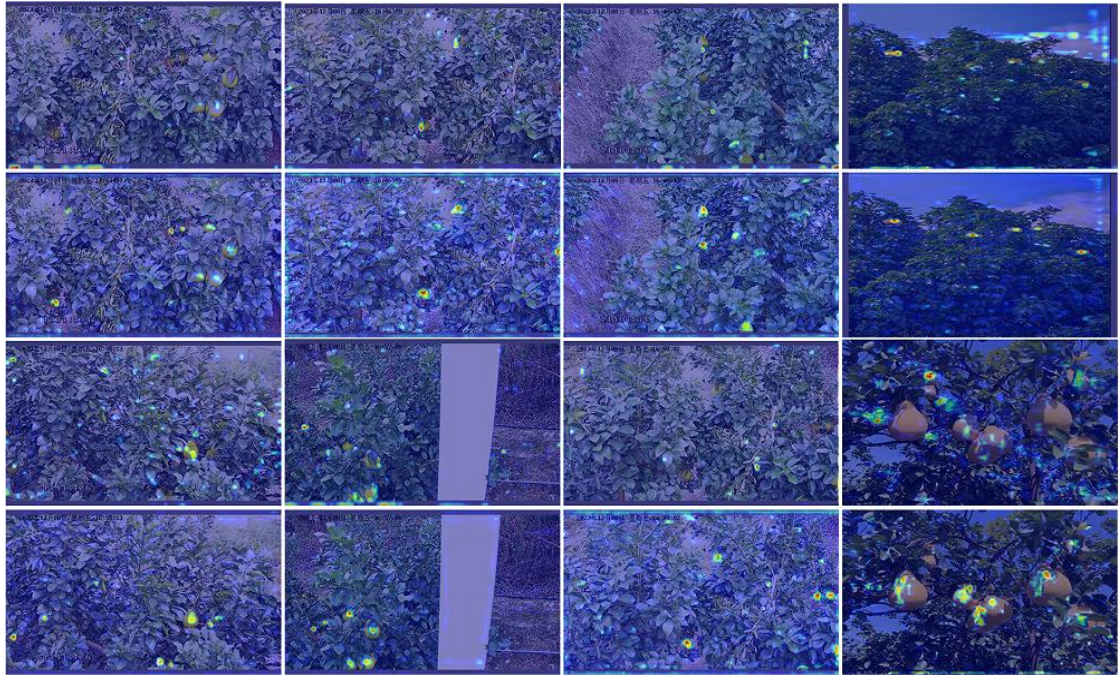

**Fig. 10.** Heat map of results.

The visualization results of the proposed SDE-DET vividly showcase its superior network performance, especially in enhancing the detection confidence. As can be seen from the figure, the first and third rows present the results of the baseline model, while the second and fourth rows display those of SDE-DET. By comparing the two groups, it is evident that SDE-DET demonstrates remarkable effectiveness in accurately detecting Shatian pomelo. The detection results of SDE-DET are more precise and confident, which confirms the robustness and reliability of this model for the specific task of Shatian pomelo detection.

**5. Discussion**

In this study, we present a new network SDE-DET, specifically designed to address the challenges in detecting Shatian pomelo in complex orchard environments. Based on Yolov8n, the model incorporates enhancements in Starnet Block, Deformable Attention, and Efficient Multi-Scale Attention mechanisms. These improvements have led to significant increases in precision, recall, mAP@0.5, mAP@0.5:0.95, and F1-score. Furthermore, to assess its practical application in Shatian pomelo harvesting, we evaluated the model parameter counts, GFLOPs, and the model size. The results show that our model demonstrates exceptional performance, with 3.29 million parameters, 32.4 GFLOPS, and a small size of 6.69 MB. These findings suggest that SDE-DET is highly suitable for embedded system implementation.

Despite our model achieving a precise and efficient performance, the issues of missed and false detections have not been fully resolved. To address these issues, future research will focus on incorporating a more diverse range of training data to better represent various orchard environments. Besides, incorporating more advanced techniques, such as reinforcement learning or self-supervised learning algorithms, can further enhance the robustness and adaptability of the model.

In our future work, we aim to continually refine our model to enhance its generalization capability and reduce reliance on specific fruit characteristics, thereby enabling the system to be adapted for detecting other types of fruits. In addition, we plan to develop systems for real-world applications, covering not only pomelo detection, but also automated processes such as picking, fruit grading, and yield prediction for

Shatian pomelo. These systems will offer farmers data-driven tools to streamline the production cycle, promote sustainable farming, and improve agricultural efficiency.

## 6. Conclusion

We construct a Shatian pomelo dataset by collecting images from field-planting bases and expanding it using various data augmentation methods to enhance the model's robustness and generalizability. Using this self-constructed dataset and the Yolov8n model as a foundation, this study proposes a deep learning model for Shatian pomelo detection, named SDE-DET. The SDE-DET model achieves precise and efficient detection of Shatian pomelo, demonstrating strong performance in complex environments.

By integrating the Star Block into the Yolov8n model, the SDE-DET model enhances its ability to extract high-dimensional feature information. The Deformable Attention improves the model's capability to detect occluded Shatian pomelos by allowing it to selectively attend to critical regions, thereby improving overall recognition accuracy. Incorporating multiple Mechanisms further enhances the model's performance, especially in detecting small Shatian pomelos. Ablation experiments demonstrate that our method improves precision, recall, mAP@0.5, mAP@0.5:0.95, and F1 values by 2.1%, 7.7%, 5.1%, 3.9%, and 5.0%, respectively, compared to the original Yolov8n. In addition, the SDE-DET model improves by 3.4%, 7.3%, 9.8%, and 5.1% over the Yolov9m, Yolov10n, DDQ-4scale and RT-DETR in mAP@0.5, respectively.

The SDE-DET model developed in this study provides a robust theoretical framework for Shatian pomelo detection. Furthermore, it offers critical technical support for the development of automated and intelligent harvesting systems, thereby facilitating improved operational efficiency and precision in agricultural production.

**Funding**

This work was supported in part by the National Natural Science Foundation of China (62462027, 62271266, 62177013 and 62277012), the National Key Research and Development Program of China (2024YFE0102100), the Scientific Research Fund Project of Hainan University (KYQD(ZR)-23113) and the Postdoctoral Research Program of Jiangsu province (2019K287), College Student Innovation Training Program Project (202410589043)

**Author Contributions**

Conceptualization, Xiaodong Bai, Aiping Yang and Jianguo Yao; Methodology, Yihao Hu, Pan Wang and Xiaodong Bai; Validation, Yihao Hu, Pan Wang, Shijie Cai and Xiaodong Bai; Formal Analysis and Investigation, Xiaodong Bai, Aiping Yang and Huazhong Liu; Resources, Aiping Yang, Xiangxiang Li and Meiping Ding; Writing and Writing-Review, Yihao Hu, Pan Wang, Shijie Cai, Hang Wang and Xiaodong Bai; Editing and Visualization, all authors. Project Administration and Funding Acquisition, Xiaodong Bai, Hongyan Liu and Jianguo Yao. All authors have read and agreed to the published version of the manuscript.

**Declaration of Competing Interest**

The authors declare that they have no known competing financial interests or personal relationships that could have appeared to influence the work reported in this paper.

**Data availability**

The data will be made available upon request.

**References**

[1] J. Yin, X. Hu, Y. Hou, S. Liu, S. Jia, C. Gan, *et al.*, "Comparative analysis of chemical compositions and antioxidant activities of different pomelo varieties from China," *Food Chemistry Advances,* vol. 2, 2023.

[2] V. Mendez, A. Perez-Romero, R. Sola-Guirado, A. Miranda-Fuentes, F. Manzano-Agugliaro, A. Zapata-Sierra, *et al.*, "In-Field Estimation of Orange Number and Size by 3D Laser Scanning," *Agronomy-Basel,* vol. 9, 2019.

[3] S. Cubero, W. S. Lee, N. Aleixos, F. Albert, and J. Blasco, "Automated Systems Based on Machine Vision for Inspecting Citrus Fruits from the Field to Postharvest-a Review," *Food and Bioprocess Technology,* vol. 9, pp. 1623-1639, 2016.

[4] H. Mirhaji, M. Soleymani, A. Asakereh, and S. Abdanan Mehdizadeh, "Fruit detection and load estimation of an orange orchard using the YOLO models through simple approaches in different imaging and illumination conditions," *Computers and Electronics in Agriculture,* vol. 191, 2021.

[5] A. Koirala, K. B. Walsh, Z. Wang, and C. McCarthy, "Deep learning for real-time fruit detection and orchard fruit load estimation: benchmarking of 'MangoYOLO'," *Precision Agriculture,* vol. 20, pp. 1107-1135, 2019.

[6] Y. Tian, G. Yang, Z. Wang, H. Wang, E. Li, and Z. Liang, "Apple detection during different growth stages in orchards using the improved YOLO-V3 model," *Computers and Electronics in Agriculture,* vol. 157, pp. 417-426, 2019.

[7] Z. Xu, J. Liu, J. Wang, L. Cai, Y. Jin, S. Zhao, *et al.*, "Realtime Picking Point Decision Algorithm of Trellis Grape for High-Speed Robotic Cut-and-Catch Harvesting," *Agronomy,* vol. 13, p. 1618, 2023.

[8] W. Zhang, C. Zheng, C. Wang, and W. Guo, "DomAda-FruitDet: Domain-Adaptive Anchor-Free Fruit Detection Model for Auto Labeling," *Plant Phenomics,* vol. 6, p. 0135, 2024.

[9] J. Rakun, D. Stajnko, and D. Zazula, "Detecting fruits in natural scenes by using spatial-frequency based texture analysis and multiview geometry," *Computers and Electronics in Agriculture,* vol. 76, pp. 80-88, 2011.

[10] F. García, J. Cervantes, A. López, and M. Alvarado, "Fruit Classification by Extracting Color Chromaticity, Shape and Texture Features: Towards an Application for Supermarkets," *IEEE Latin America Transactions,* vol. 14, pp. 3434-3443, 2016.

[11] F. Xiao, H. Wang, Y. Li, Y. Cao, X. Lv, and G. Xu, "Object Detection and Recognition Techniques Based on Digital Image Processing and Traditional Machine Learning for Fruit and Vegetable Harvesting Robots: An Overview and Review," *Agronomy-Basel,* vol. 13, 2023.

[12] X. Bai, P. Liu, Z. Cao, H. Lu, H. Xiong, A. Yang, *et al.*, "Rice Plant Counting, Locating, and Sizing Method Based on High-Throughput UAV RGB Images," *Plant Phenomics,* vol. 5, 2023.

[13] H. Lu, Z. Cao, Y. Xiao, B. Zhuang, and C. Shen, "TasselNet: counting maize tassels in the wild via local counts regression network," *Plant Methods,* vol. 13, 2017.

[14] J. Liu, Y. Zhao, W. Jia, and Z. Ji, "DLNet: Accurate segmentation of green fruit in obscured environments," *Journal of King Saud University - Computer and Information Sciences,* vol. 34, pp. 7259-7270, 2022.

[15] W. Jia, Z. Zhang, W. Shao, S. Hou, Z. Ji, G. Liu*, et al.*, "FoveaMask: A fast and accurate deep learning model for green fruit instance segmentation," *Computers and Electronics in Agriculture,* vol. 191, p. 106488, 2021.

[16] Q. Wang, X. Fan, Z. Zhuang, T. Tjahjadi, S. Jin, H. Huan*, et al.*, "One to All: Toward a Unified Model for Counting Cereal Crop Heads Based on Few-Shot Learning," *Plant Phenomics,* vol. 6, p. 0271, 2024.

[17] Y. Zhao, S. Wang, Q. Zeng, W. Ni, H. Duan, N. Xie*, et al.*, "Informed-Learning-Guided Visual Question Answering Model of Crop Disease," *Plant Phenomics,* vol. 6, p. 0277, 2024.

[18] J. Chen, Q. Li, and D. Jiang, "From Images to Loci: Applying 3D Deep Learning to Enable Multivariate and Multitemporal Digital Phenotyping and Mapping the Genetics Underlying Nitrogen Use Efficiency in Wheat," *Plant Phenomics,* vol. 6, p. 0270, 2024.

[19] S. Ren, K. He, R. Girshick, and J. Sun, "Faster R-CNN: Towards Real-Time Object Detection with Region Proposal Networks," *IEEE Transactions on Pattern Analysis and Machine Intelligence,* vol. 39, pp. 1137-1149, 2017.

[20] Z. Cai and N. Vasconcelos, "Cascade R-CNN: Delving Into High Quality Object Detection," in *Proceedings of the IEEE/CVF Conference on Computer Vision and Pattern Recognition (CVPR)*, 2018, pp. 6154–6162.

[21] W. Liu, D. Anguelov, D. Erhan, C. Szegedy, S. Reed, C.-Y. Fu*, et al.*, "SSD: Single Shot MultiBox Detector," *Arxiv,* 2016.

[22] J. Redmon and A. Farhadi, "YOLO9000: Better, Faster, Stronger," *Arxiv,* 2016.

[23] J. Redmon and A. Farhadi, "YOLOv3: An Incremental Improvement," *Arxiv,* 2018.

[24] J. Redmon, S. Divvala, R. Girshick, and A. Farhadi, "You Only Look Once: Unified, Real-Time Object Detection," *Arxiv,* 2016.

[25] K. Duan, S. Bai, L. Xie, H. Qi, Q. Huang, and Q. Tian, "CenterNet: Keypoint Triplets for Object Detection," in *IEEE/CVF International Conference on Computer Vision (ICCV)*, 2019, pp. 6568-6577.

[26] M. Tan, R. Pang, and Q. V. Le, "EfficientDet: Scalable and Efficient Object Detection," in *Proceedings of the IEEE/CVF Conference on Computer Vision and Pattern Recognition (CVPR)*, 2020, pp. 10778-10787.

[27] W. Jia, Y. Tian, R. Luo, Z. Zhang, J. Lian, and Y. Zheng, "Detection and segmentation of overlapped fruits based on optimized mask R-CNN application in apple harvesting robot," *Computers and Electronics in Agriculture,* vol. 172, 2020.

[28] F. Gao, L. Fu, X. Zhang, Y. Majeed, R. Li, M. Karkee, *et al.*, "Multi-class fruit-on-plant detection for apple in SNAP system using Faster R-CNN," *Computers and Electronics in Agriculture,* vol. 176, 2020.

[29] J. D. López-Barrios, J. A. Escobedo Cabello, A. Gómez-Espinosa, and L.-E. Montoya-Cavero, "Green Sweet Pepper Fruit and Peduncle Detection Using Mask R-CNN in Greenhouses," *Applied Sciences,* vol. 13, p. 6296, 2023.

[30] K. Goyal, P. Kumar, and K. Verma, "AI-based fruit identification and quality detection system," *Multimedia Tools and Applications,* vol. 82, pp. 24573-24604, 2022.

[31] S. Yang, W. Wang, S. Gao, and Z. Deng, "Strawberry ripeness detection based on YOLOv8 algorithm fused with LW-Swin Transformer," *Computers and Electronics in Agriculture,* vol. 215, 2023.

[32] W. Chen, M. Liu, C. Zhao, X. Li, and Y. Wang, "MTD-YOLO: Multi-task deep convolutional neural network for cherry tomato fruit bunch maturity detection," *Computers and Electronics in Agriculture,* vol. 216, 2024.

[33] A. Zhu, R. Zhang, L. Zhang, T. Yi, L. Wang, D. Zhang, *et al.*, "YOLOv5s-CEDB: A robust and efficiency Camellia oleifera fruit detection algorithm in complex natural scenes," *Computers and Electronics in Agriculture,* vol. 221, 2024.

[34] J. Lu, P. Chen, C. Yu, Y. Lan, L. Yu, R. Yang, *et al.*, "Lightweight green citrus fruit detection method for practical environmental applications," *Computers and Electronics in Agriculture,* vol. 215, 2023.

[35] C. Yu, X. Shi, W. Luo, J. Feng, Z. Zheng, A. Yorozu, *et al.*, "MLG-YOLO: A Model for Real-Time Accurate Detection and Localization of Winter Jujube in Complex Structured Orchard Environments," *Plant Phenomics,* vol. 6, p. 0258, 2024.

[36] Z. Xia, X. Pan, S. Song, L. E. Li, and G. Huang, "Vision Transformer with Deformable Attention," in *Proceedings of the IEEE/CVF Conference on Computer Vision and Pattern Recognition (CVPR)*, 2022, pp. 4794-4803.

[37] D. Ouyang, S. He, G. Zhang, M. Luo, H. Guo, J. Zhan, *et al.*, "Efficient Multi-Scale Attention Module with Cross-Spatial Learning," in *IEEE International Conference on Acoustics, Speech and Signal Processing (ICASSP)*, 2023, pp. 1-5.

[38] C.-Y. Wang, I. H. Yeh, and H.-Y. M. Liao, "YOLOv9: Learning What You Want to Learn Using Programmable Gradient Information," *ArXiv,* 2024.

[39] A. Wang, H. Chen, L. Liu, K. Chen, Z. Lin, J. Han, *et al.*, "YOLOv10: Real-Time End-to-End Object Detection," *Arxiv,* 2024.

[40] C. Lyu, W. Zhang, H. Huang, Y. Zhou, Y. Wang, Y. Liu, *et al.*, "RTMDet: An Empirical Study of Designing Real-Time Object Detectors," *Arxiv,* 2022.

[41] H. Zhang, F. Li, S. Liu, L. Zhang, H. Su, J. Zhu, *et al.*, "DINO: DETR with Improved DeNoising Anchor Boxes for End-to-End Object Detection," in *International Conference on Learning Representations (ICLR)*, 2023, pp. 13619-13627.

[42] S. Zhang, X. Wang, J. Wang, J. Pang, C. Lyu, W. Zhang, *et al.*, "Dense Distinct Query for End-to-End Object Detection," in *Proceedings of the IEEE/CVF Conference on Computer Vision and Pattern Recognition (CVPR)*, 2023, pp. 7329-7338.

[43] Y. Zhao, W. Lv, S. Xu, J. Wei, G. Wang, Q. Dang, *et al.*, "DETRs Beat YOLOs on Real-time Object Detection," *Arxiv,* 2024.